%% file: main.tex
\documentclass[journal]{IEEEtran}

\usepackage{xcolor}
\usepackage{enumitem}
\usepackage{float}

\usepackage[normalem]{ulem}

\input{setup/packages}
\input{setup/math}

\input{setup/defs}

\hypersetup{
  pdftitle={NAT: Scaling Causal Neural Action Tokenizers Across Embodiments},
}

\begin{document}
\bstctlcite{IEEEexample:BSTcontrol}

\title{NAT: Scaling Causal Neural Action Tokenizers Across Embodiments}

\author{Ahad Jawaid, Alexander Swerdlow, and Yu Xiang%
  \thanks{A. Swerdlow is with Stanford University. Y. Xiang is with The University of Texas at Dallas.}}

\markboth{}{}

\maketitle

\input{content/abstract}

\begin{IEEEkeywords}
Robot learning, action tokenization, autoregressive policies, vision-language-action models, robot manipulation
\end{IEEEkeywords}

\input{content/body}

\appendices
\input{content/appendix}

\section*{Acknowledgment}
We thank MicroAGI and DeepReach AI for providing access to their human hand-motion data, and Chaoqi Liu for helpful early conversations.

\bibliographystyle{IEEEtran}
\bibliography{main}

\end{document}

%% file: setup/packages.tex
\usepackage{algorithm}
\usepackage{algorithmicx}
\usepackage{algcompatible}
\usepackage[noEnd,commentColor=black]{algpseudocodex}

\usepackage{amsmath}
\usepackage{amssymb}
\usepackage{mathtools}
\usepackage{amsthm}
\usepackage{cancel}

\usepackage{thmtools} 
\usepackage{thm-restate}

\usepackage{hyperref}       
\usepackage[capitalize,noabbrev]{cleveref}

\crefname{assumption}{Assumption}{Assumption}
\theoremstyle{plain}

\usepackage[textsize=tiny]{todonotes}

\usepackage[utf8]{inputenc} 
\usepackage[T1]{fontenc}    

\usepackage{url}            
\usepackage{booktabs}       

\usepackage{xcolor} 
\usepackage{graphicx}
\usepackage{blindtext}
\usepackage{etoolbox}
\usepackage[export]{adjustbox}

\usepackage{wrapfig}
\usepackage{pifont}

\usepackage{amsfonts}       
\usepackage{nicefrac}       
\usepackage[nopatch=showhyphens]{microtype} 

\usepackage{multirow}
\usepackage{multicol}
\usepackage{soul}
\usepackage{tikz-cd}

\usepackage{transparent}

\usepackage{tcolorbox}
\tcbuselibrary{theorems}

\definecolor{ourcolor}{HTML}{00A89D}
\definecolor{ourpurple}{HTML}{CB297B}
\definecolor{ourblue}{HTML}{0076BA}
\definecolor{ourmiddle}{HTML}{66509B}
\definecolor{ourlightblue}{HTML}{F5FAFE}
\definecolor{ourlightmiddle}{HTML}{F7F6F9}

\declaretheoremstyle[
    headfont=\bfseries,
    bodyfont=\normalfont,
    spaceabove=10pt, spacebelow=10pt,
    headpunct={},
    postheadspace=1em,
    mdframed={
        backgroundcolor=ourcolor!5,
        linecolor=ourcolor!50!black,
        linewidth=1pt,
        roundcorner=4pt
    }
]{colorboxed}

\declaretheoremstyle[
    headfont=\bfseries,
    bodyfont=\normalfont,
    spaceabove=10pt, spacebelow=10pt,
    headpunct={},
    postheadspace=1em,
    mdframed={
        backgroundcolor=ourblue!5,
        linecolor=ourblue!50!black,
        linewidth=1pt,
        roundcorner=4pt
    }
]{bluecolorboxed}

\declaretheoremstyle[
    headfont=\bfseries,
    bodyfont=\normalfont,
    spaceabove=10pt, spacebelow=10pt,
    headpunct={},
    postheadspace=1em,
    mdframed={
        backgroundcolor=ourpurple!5,
        linecolor=ourpurple!50!black,
        linewidth=1pt,
        roundcorner=4pt
    }
]{purplecolorboxed}

\declaretheoremstyle[
    headfont=\bfseries,
    bodyfont=\normalfont,
    spaceabove=10pt, spacebelow=10pt,
    headpunct={},
    postheadspace=1em,
    mdframed={
        backgroundcolor=black!5,
        linecolor=black!50!black,
        linewidth=1pt,
        roundcorner=4pt
    }
]{greycolorboxed}

\usepackage{tikz}
\usetikzlibrary{decorations.pathreplacing,calc}

\usepackage{capt-of}

%% file: setup/math.tex
\usepackage{amsmath,amsfonts,bm, stackengine}

\DeclarePairedDelimiterX{\infdivx}[2]{(}{)}{%
  #1\;\delimsize\|\;#2%
}

\def\eqref#1{equation~\ref{#1}}

\def\1{\bm{1}}

\DeclareMathAlphabet{\mathsfit}{\encodingdefault}{\sfdefault}{m}{sl}
\SetMathAlphabet{\mathsfit}{bold}{\encodingdefault}{\sfdefault}{bx}{n}



%% file: setup/defs.tex
\definecolor{darkblue}{rgb}{0,0.08,0.45}
\definecolor{cred}{HTML}{D62728}
\definecolor{cblue}{HTML}{1F77B4}
\definecolor{cgreen}{HTML}{79AB76}
\definecolor{cgrey}{rgb}{0.6,0.6,0.6}

\renewcommand{\mathbf}{\boldsymbol}

\makeatletter

\usepackage{threeparttable}
\newcommand{\tabnote}[1]{\par\vspace{3pt}{\footnotesize\noindent\raggedright #1\par}}

%% file: content/abstract.tex
\begin{abstract}
Action tokenizers let robot policies treat control as autoregressive next-token prediction. As these policies are trained on larger and more diverse datasets, they will need a universal tokenizer that can encode actions from many embodiments and data sources, including ones it has not seen. We study action tokenization at scale, training on 49k hours of action data from real-world robots, handheld grippers, simulation, and humans. We introduce Neural Action Tokenizer (NAT), a causal tokenizer whose decoder lets actions be executed as their tokens are generated. Across four simulation benchmarks, NAT outperforms FAST and prior neural action tokenizers when tokenizing unseen robot embodiments, and in vision-language-action models it outperforms FAST+ across four benchmarks and on a real robot when acting through its tokens. NAT tokens also close the performance gap to a flow-matching action head on some benchmarks while lowering the inference latency for first actions. Our experiments also find that robot data may not be necessary to learn an effective action space: a NAT trained only on human hand motion performs close to tokenizers trained on robot data.
\end{abstract}

%% file: content/body.tex
\section{Introduction}

Large language models \cite{minaee2024large} have benefited from a simple scalable training recipe: convert text into a shared vocabulary of tokens, then train a single autoregressive model on a large corpus of tokens. A similar recipe may help scale robotic learning. Robot data, however, comes from many embodiments, often with different action spaces, control frequencies, and ranges of motion, and is collected in different ways: teleoperation on real robots, handheld grippers such as UMI \cite{chi2024umi}, simulation, and recordings of human hands. A foundation robot policy that uses this recipe will likely need a tokenizer that can encode large and diverse sets of actions, ideally including those from embodiments it has not seen.

The action tokenizer therefore plays an important role in training autoregressive robot policies, and its design can substantially affect policy performance, training efficiency, and inference speed \cite{pertsch2025fast}. Early autoregressive policies discretized each action dimension into uniform bins~\cite{pmlr-rt2-zitkovich23a, openvla2024}. Binning can be applied to most robots, but it tends to produce long token sequences that slow down inference. FAST~\cite{pertsch2025fast} compresses action chunks with a frequency-domain transform and byte pair encoding, yielding shorter sequences and faster training, and the released FAST+~\cite{pertsch2025fast} is trained on 1M robot trajectories such that it can work across embodiments. A growing line of work instead learns the tokenizer: a neural encoder, a quantizer~\cite{lee2022rvq, mentzer2024fsq}, and a decoder that reconstructs the action chunk~\cite{wang2025vqvla, liu2026faster, liu2026oat, jawaid2026nac, dong2026actioncodec}. These tokenizers can represent actions with fewer tokens than FAST~\cite{pertsch2025fast}, but most are trained on comparatively small datasets, so how they behave on the scale and diversity of data a foundation policy would require remains largely unexplored.

In this work, we study action tokenization at scale. We assemble a corpus of 49k hours of action data from 156 datasets spanning real-world robots, universal manipulation interfaces (UMIs) \cite{chi2024umi}, simulated robots, and human hand motion, and use it to train and evaluate action tokenizers. We also introduce the Neural Action Tokenizer (NAT)\footnote{Website: \url{https://www.ahadjawaid.com/nat}. Tokenizer: \url{https://huggingface.co/ahad-j/nat-tokenizer-h32-l4-c6400}.}, a causal neural tokenizer that encodes action chunks into short token sequences using finite scalar quantization (FSQ)~\cite{mentzer2024fsq} at multiple temporal scales. Its causal decoder supports streaming: actions can be decoded as their tokens are generated, so the robot can begin executing while the policy is still predicting the rest of the chunk, and the policy can condition its next chunk on the actions still executing. We then pretrain vision-language-action models (VLAs) with NAT to test whether these results carry over to large pretrained policies, and compare acting through NAT tokens with acting through a flow-matching action head~\cite{lipman2023flow, black2024pi0}.

Our contributions are:
\begin{itemize}[leftmargin=*, labelindent=1em]
  \item \textbf{State-of-the-art neural action tokenization.} NAT outperforms FAST~\cite{pertsch2025fast} and prior neural action tokenizers (OAT~\cite{liu2026oat}, ActionCodec~\cite{dong2026actioncodec}, and NAC~\cite{jawaid2026nac}) in mean success on unseen robot embodiments.
  \item \textbf{A universal streaming action tokenizer.} We train NAT on 49k hours of actions from real-world robots, handheld grippers, simulation, and humans. Its causal decoder supports immediate execution, and its time-ordered tokens let a policy condition its next chunk on the actions still executing without retraining.
  \item \textbf{Human data for action tokenizers.} Our experiments suggest that robot data may not be necessary to train an action tokenizer: a NAT trained only on human hand motion is nearly as effective as one trained on all sources.
  \item \textbf{Pretraining vision-language-action models with NAT tokens.} We show that NAT tokens can help pretrain vision-language-action models: further training a pretrained vision-language model on 1.8k hours of robot data through NAT tokens, before fine-tuning, raises mean success on four simulation benchmarks over fine-tuning the vision-language model directly.
  \item \textbf{Flow action heads and action tokens.} Acting through NAT tokens produces first actions sooner than acting through a flow-matching action head, and on some benchmarks matches its success: on LIBERO~\cite{liu2023libero}, the two are within the 95\% intervals.
\end{itemize}

\section{Related Work}

\subsection{Autoregressive Robot Policies}

Robot foundation models are commonly built by fine-tuning a pretrained vision-language model on robot data~\cite{black2024pi0, black2025pi0.5, nvidia2025groot, fang2026molmoact2actionreasoningmodels}. Since these backbones are autoregressive transformers~\cite{vaswani2017attention}, actions can be tokenized and predicted with next-token prediction; early work showed that even simple binning transfers web-scale knowledge to robot control~\cite{pmlr-rt2-zitkovich23a, openvla2024}. Other models, such as $\pi_0$~\cite{black2024pi0} and GR00T~\cite{nvidia2025groot}, instead decode actions with a flow-matching or diffusion head which showed improvements in inference latency. FAST~\cite{pertsch2025fast} showed that mid-training on discrete action tokens improves training efficiency and policy performance, and later work adopts this for models with a continuous action expert~\cite{black2025pi0.5, driess2026knowledge, fang2026molmoact2actionreasoningmodels}.

\begin{figure*}[t]
  \centering
  \includegraphics[width=\textwidth]{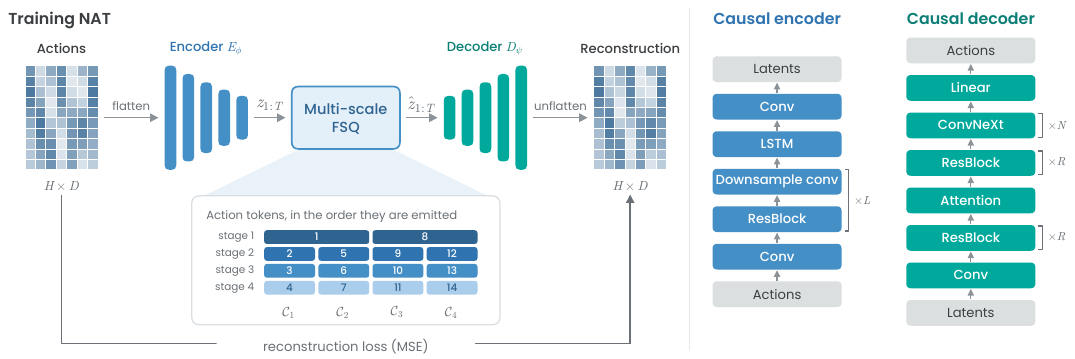}
  \caption{\textbf{NAT architecture.} \emph{Left:} an $H \times D$ action chunk is flattened, encoded into latents $z_{1:T}$, quantized by multi-scale FSQ into coarse-to-fine tokens, and decoded back; NAT is trained end to end with a mean-squared reconstruction loss. The inset shows the 14 tokens of one chunk of the universal NAT ($T=4$ latents, stage spans $[2,1,1,1]$), numbered in the order they are emitted; $\mathcal{C}_t$ holds the tokens of latent $t$. \emph{Right:} the causal encoder and decoder layer stacks, read bottom to top; $L$, $R$, and $N$ denote repeated blocks (Appendix~\ref{app:hyperparameters}).}
  \label{fig:architecture}
\end{figure*}

\subsection{Action Tokenizers}

Binning each action dimension~\cite{pmlr-rt2-zitkovich23a, openvla2024} is simple but produces long token sequences that can make real-time inference impractical. FAST~\cite{pertsch2025fast} uses handcrafted frequency-based compression to produce much shorter sequences, improving training efficiency and policy performance, and its variant FAST+ \cite{pertsch2025fast}, fitted on 1M real robot action sequences, tokenizes new action spaces zero-shot. Learned tokenizers compress actions further with vector-quantized autoencoders~\cite{wang2025vqvla}, and have been extended to multiple embodiments~\cite{liu2026faster}, aligned with language~\cite{dong2026actioncodec}, ordered so that partial token sequences can be decoded~\cite{liu2026oat}, and scaled to 2B action frames with vision-language alignment~\cite{kang2026xtokenizermultimodalactiontokenizer}. Most neural tokenizers have been trained on comparatively smaller datasets, so it remains largely unexplored how learned tokenizers behave when trained on large, heterogeneous action data spanning real robots, UMIs \cite{chi2024umi}, simulation, and humans.

\subsection{Real-Time Execution}

Action chunking~\cite{zhao-act, chi2025diffusion} lets a policy predict several future actions per inference, but the robot must wait for each new chunk, and switching between independently sampled chunks can cause discontinuities. Many VLAs predict chunks with a diffusion or flow-matching action head~\cite{black2024pi0, nvidia2025groot}, which generates every action of a chunk in parallel over a fixed number of denoising steps; knowledge insulation~\cite{driess2026knowledge} trains such a head beside discrete action tokens, and consistency distillation~\cite{prasad2024consistency} reduces the number of denoising steps. Other work hides inference latency during execution: asynchronous inference~\cite{shukor2025smolvla} predicts the next chunk while the current one runs, bidirectional decoding~\cite{liu2024bid} selects among sampled chunks for coherence with the previous one, and real-time chunking~\cite{black2026rtc} inpaints the actions still executing into a flow policy's sampling so that consecutive chunks join smoothly. Autoregressive policies, in contrast, usually decode a chunk only after all of its tokens are generated~\cite{pertsch2025fast}. NAT's time-ordered tokens and causal decoder give them both properties: actions are executed as their tokens are generated, and executing actions are fixed as a token prefix.

\section{Preliminaries}

\textbf{Action chunking.} We consider policies that, given an observation $o$ (camera images, proprioception, and a task), predict a chunk of $H$ future actions $a_{1:H} \in \mathbb{R}^{H \times D}$, where $D$ is the action dimension. {} During execution, the robot executes the first $s \le H$ actions of each chunk before querying the policy again. We refer to $s$ as the execution horizon~\cite{black2026rtc}.

\textbf{Action tokenization.} Autoregressive policies that predict discrete tokens, the setting we study, require continuous action chunks to first be converted into a token sequence. A tokenizer $\mathcal{T}$ maps an action chunk to a sequence of $L$ tokens, $\mathcal{T}(a_{1:H}) = c_{1:L}$ with $c_i \in \mathcal{V}, i=1,\ldots,L$, where $\mathcal{V}$ is the tokenizer's vocabulary, and a detokenizer $\mathcal{T}^{-1}$ maps a token sequence back to an action chunk, $\mathcal{T}^{-1}(c_{1:L}) = \hat{a}_{1:H}$. The policy only predicts in the token space, while the robot executes the continuous action reconstructed by $\mathcal{T}^{-1}$.

Existing tokenizers differ in how they construct this mapping. \emph{Binning}~\cite{pmlr-rt2-zitkovich23a, openvla2024} normalizes each action dimension to $[-1, 1]$ and assigns each value to one of $N$ uniform bins, giving one token per dimension per timestep ($L = HD$). FAST ~\cite{pertsch2025fast} applies the discrete cosine transform \cite{ahmed-dct} to each action dimension, rounds the resulting coefficients, and compresses them with byte-pair encoding \cite{gage1994bpe}. \emph{Neural tokenizers}~\cite{wang2025vqvla, liu2026faster, liu2026oat, jawaid2026nac, dong2026actioncodec} instead learn the mapping: an encoder compresses the chunk into a short sequence of latent vectors, a quantizer~\cite{lee2022rvq, mentzer2024fsq} maps each latent to an entry of a discrete codebook, and a decoder reconstructs the chunk.

\textbf{Streaming.} For most tokenizers, the detokenizer $\mathcal{T}^{-1}$ needs the complete token sequence before it can decode any action, so the robot must wait for the policy to generate all $L$ tokens. We call a tokenizer \emph{streaming} if every prefix of the token sequence already determines a prefix of the action chunk, and those actions do not change as further tokens are generated. With a streaming tokenizer, the robot can begin executing a chunk while the policy is still generating it.

\section{Neural Action Tokenizer}

We introduce the Neural Action Tokenizer (NAT), a neural tokenizer whose decoder enables streaming.

\subsection{Architecture}
\label{sec:architecture}

NAT builds on the Neural Action Codec (NAC)~\cite{jawaid2026nac}, an action tokenizer adapted from neural audio codecs~\cite{zeghidour2021soundstream, defossez2023encodec}. We start from the variant of NAC trained without adversarial losses and build on it with three changes, each guided by an ablation (Section~\ref{sec:ablations}). First, we replace NAC's multi-scale residual vector quantization (RVQ) with a multi-scale variant of finite scalar quantization (FSQ)~\cite{mentzer2024fsq}, which substantially improved downstream policy performance. Second, we replace NAC's inverse short-time Fourier transform (ISTFT) output head with a linear layer, which performed comparably with a simpler design. Third, we make the encoder and decoder causal, which is what enables streaming. This caused only a small change in policy performance relative to a non-causal NAT. Figure~\ref{fig:architecture} shows the resulting model, which consists of an encoder $E_\phi$, a multi-scale FSQ quantizer, and a decoder $D_\psi$.

\textbf{Encoder.} As in NAC~\cite{jawaid2026nac}, we flatten the action chunk time-major into a one-dimensional signal $w = [a_{1,1}, \dots, a_{1,D}, a_{2,1}, \dots, a_{H,D}] \in \mathbb{R}^{HD}$, so that each action occupies a contiguous segment. The encoder is a SEANet network~\cite{tagliasacchi20_interspeech, zeghidour2021soundstream}: a stack of strided one-dimensional convolutions with residual blocks, followed by an LSTM as in EnCodec~\cite{defossez2023encodec}. It downsamples $w$ into $T$ latent vectors $z_{1:T} = E_\phi(w)$, each summarizing $k = H/T$ consecutive actions. To make the encoder causal, we pad convolutions only on the left and run the LSTM forward, so latent $z_t$ depends only on the actions $a_{1:tk}$.

\textbf{Multi-scale FSQ.} FSQ~\cite{mentzer2024fsq} quantizes a vector without a learned codebook by bounding each of its $d$ channels and rounding it to one of $\ell_j$ levels,
\begin{equation}
\mathrm{FSQ}(x)_j = \mathrm{round}\big(h_j \tanh(x_j + \beta_j) - o_j\big), \qquad j = 1, \dots, d,
\end{equation}
where $h_j = (\ell_j - 1)/2$; for even $\ell_j$, $o_j = 1/2$ and $\beta_j = \operatorname{artanh}(o_j / h_j)$ give $\ell_j$ levels (else $o_j = \beta_j = 0$).
Each rounded vector is one of $V = \prod_j \ell_j$ codes, and its index is the token. During training, gradients are passed through the rounding unchanged (the straight-through estimator).

We apply FSQ in $M$ stages, following residual quantization~\cite{lee2022rvq}: each stage quantizes the part of $z$ that the previous stages have not yet captured, called the residual. Following SNAC~\cite{siuzdak2024snac}, each stage also works at its own time resolution: stage $m$ averages every $p_m$ consecutive latents into one vector before quantizing, so it produces one token for every $p_m$ latents. Starting from $r_1 = z_{1:T}$, stage $m$ averages the residual, quantizes it, and subtracts the result:
\begin{align}
\hat{q}_m &= W^{(m)}_{\text{out}} \, \mathrm{FSQ}\big(W^{(m)}_{\text{in}} \, \mathrm{Avg}_{p_m}(r_m)\big), \label{eq:stage} \\
r_{m+1} &= r_m - \mathrm{Repeat}_{p_m}(\hat{q}_m),
\end{align}
where $\mathrm{Avg}_{p}$ averages each window of $p$ latents, $\mathrm{Repeat}_{p}$ copies each vector $p$ times to return to one vector per latent, and $W^{(m)}_{\text{in}}, W^{(m)}_{\text{out}}$ are learned linear projections. The quantized latents are the sum of all stages,
\begin{equation}
\hat{z}_{1:T} = \mathrm{MFSQ}(z_{1:T}) = \sum_{m=1}^{M} \mathrm{Repeat}_{p_m}(\hat{q}_m), \label{eq:mfsq}
\end{equation}
and we write $c^{(m)}_i$ for the $i$-th token of stage $m$.

\textbf{Decoder.} The decoder follows NAC's Vocos-style backbone~\cite{siuzdak2023vocos}, a stack of convolutional, self-attention, and ConvNeXt~\cite{liu2022convnet} blocks operating on the quantized latents. We make it causal in three ways: convolutions are padded only on the left, each latent attends only to itself and a fixed window of preceding latents, and normalization statistics are computed separately for each latent rather than across time. A linear head then maps the decoder output at latent $t$ to the $k$ actions that latent represents. As a result, the reconstructed actions $\hat{a}_{(t-1)k+1:tk}$ depend only on $\hat{z}_{1:t}$.

\textbf{Training objective.} NAT is trained end to end to reconstruct its input. Writing the reconstruction as $\hat{w} = D_\psi\big(\mathrm{MFSQ}(E_\phi(w))\big)$, we minimize the mean squared error
\begin{equation}
\mathcal{L}_{\text{NAT}}(\phi, \psi) = \mathbb{E}_{a_{1:H} \sim \mathcal{D}} \Big[\tfrac{1}{HD} \lVert \hat{w} - w \rVert_2^2\Big]
\end{equation}
over a dataset $\mathcal{D}$ of action chunks.

\subsection{Tokenization and Detokenization}
\label{sec:tokenization}

\textbf{Tokenization.} To tokenize a chunk, NAT encodes it and quantizes it with multi-scale FSQ, which gives a set of tokens for every stage. We then interleave the stages into a single sequence in time order, so that tokens describing the beginning of the chunk come first. Concretely, we walk through the latents $t = 1, \dots, T$ and, at each one, emit every token that latent needs and that has not been emitted yet, from the coarsest stage to the finest. Calling these tokens $\mathcal{C}_t$, the token sequence is
\begin{equation}
\mathcal{T}(a_{1:H}) = [\,\mathcal{C}_1, \mathcal{C}_2, \dots, \mathcal{C}_T\,].
\end{equation}
For example, with four stages where the first produces one token per two latents and the rest one token per latent, odd latents contribute four tokens (one per stage) and even latents contribute three, since their first-stage token was already emitted with the previous latent. Since recovering a latent requires one token from every stage, the actions of latent $t$ can be decoded only after all tokens in $\mathcal{C}_t$ have been generated. Streaming therefore advances in steps of $|\mathcal{C}_t|$ tokens rather than one token at a time: in this example, the first actions are decoded after four tokens, and each later latent's actions after three or four more.

\begin{figure}[t]
  \centering
  \includegraphics[width=\columnwidth]{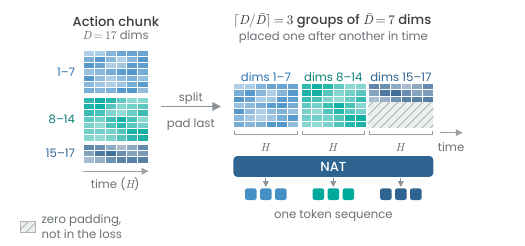}
  \caption{\textbf{Varying action dimensions.} The $D$ action dimensions are split into
  $\lceil D/\bar{D} \rceil$ groups of $\bar{D}$ slots ($\bar{D} = 7$ here and in our experiments), one per arm
  where the dataset defines arms and consecutive dimensions otherwise, and the last group is zero-padded. The
  groups are placed one after another in time, NAT encodes each as an $H \times \bar{D}$ chunk, and their
  tokens form one sequence. Padded slots are excluded from the reconstruction loss.}
  \label{fig:action-dims}
\end{figure}

\textbf{Varying action dimensions.} Robots differ in their number of action dimensions, while NAT encodes chunks of a fixed width $\bar{D}$. We therefore split each action vector into groups of at most $\bar{D}$ dimensions, one per arm where the dataset defines arms and consecutive dimensions otherwise, and zero-pad each group to $\bar{D}$. Each group is tokenized as its own $H \times \bar{D}$ chunk, and the tokens of all groups are concatenated (Figure~\ref{fig:action-dims}). Padded dimensions are excluded from the loss (Appendix~\ref{app:corpus}).

\textbf{Detokenization.} To detokenize, each token is mapped back to its FSQ code, the codes are projected and summed across stages to recover $\hat{z}_{1:T}$, and the decoder produces the action chunk. This ordering is what makes NAT streaming: the first $t$ groups $\mathcal{C}_{1:t}$ contain every token needed to compute $\hat{z}_{1:t}$, and because the decoder is causal, $\hat{z}_{1:t}$ is all it needs to produce the actions of latent $t$.

\subsection{Autoregressive Policy}
\label{sec:policy}

Following NAC~\cite{jawaid2026nac}, the policy $\pi_\theta$ is a transformer decoder~\cite{vaswani2017attention} that generates NAT tokens autoregressively while cross-attending to features of the observation. Tokens from different FSQ stages share the same code indices but carry different information, so we give each stage its own range of $V$ token ids, giving a vocabulary of $MV$ ids plus a beginning-of-sequence token. \emph{NAT is trained first and kept frozen while the policy is trained.} Given a demonstration $(o, a_{1:H})$ with token sequence $c_{1:L} = \mathcal{T}(a_{1:H})$, the policy is trained with the standard next-token prediction loss, averaged over the chunk's tokens,
\begin{equation}
\mathcal{L}_{\pi}(\theta) = -\,\mathbb{E}_{(o, a_{1:H}) \sim \mathcal{D}} \Big[ \frac{1}{L} \sum_{i=1}^{L} \log \pi_\theta\big(c_i \mid o, c_{<i}\big) \Big]. \label{eq:policy}
\end{equation}

\begin{table*}[t]
  \centering
  \caption{Comparison with prior action tokenizers}
  \label{tab:comparison}
  \footnotesize
  \setlength{\tabcolsep}{4pt}
  \begin{threeparttable}
  \begin{tabular*}{\linewidth}{@{\extracolsep{\fill}}lccccc}
    \toprule
    Method & LIBERO-Long~\cite{liu2023libero} & Meta-World~\cite{yu2020metaworld} & RoboMimic~\cite{mandlekar2022matters} & RoboCasa~\cite{nasiriany2024robocasa} & Average \\
    \midrule
    NAT (ours) & $\mathbf{74.4 \pm 3.8}$ & $38.2 \pm 6.6$ & $38.6 \pm 2.5$ & $\mathbf{36.8 \pm 3.3}$ & $\mathbf{47.0 \pm 1.5}$ \\
    FAST \cite{pertsch2025fast} & $68.6 \pm 0.7$ & $\mathbf{39.4 \pm 1.4}$ & $35.1 \pm 3.4$ & $30.8 \pm 7.9$ & $43.4 \pm 1.7$ \\
    OAT \cite{liu2026oat} & $70.3 \pm 2.9$ & $9.2 \pm 3.1$ & $\mathbf{40.0 \pm 3.7}$ & $26.8 \pm 4.7$ & $36.6 \pm 2.3$ \\
    ActionCodec \cite{dong2026actioncodec} & $61.8 \pm{3.8}$ & $32.2 \pm{2.3}$ & $39.1 \pm{1.6}$ & $30.5 \pm{4.5}$ & $40.9 \pm{1.6}$ \\
    NAC \cite{jawaid2026nac} & $57.2 \pm 1.3$ & $7.9 \pm 2.5$ & $26.1 \pm 4.2$ & $25.0 \pm 2.3$ & $29.0 \pm 1.2$ \\
    Diffusion Policy \cite{chi2025diffusion} & $57.5 \pm 4.6$ & $31.8 \pm 4.3$ & $29.2 \pm 3.5$ & $25.2 \pm 5.0$ & $35.9 \pm 2.6$ \\
    \bottomrule
  \end{tabular*}
  \tabnote{Success (\%) of one cross-embodiment policy per tokenizer, trained jointly on the four benchmarks.}
  \end{threeparttable}
\end{table*}

\textbf{Streaming decoding.} At inference, the policy generates tokens one at a time. As soon as the tokens $\mathcal{C}_t$ of a latent are complete, NAT decodes that latent's $k$ actions and passes them to the robot, while the policy continues generating the remaining tokens (Algorithm~\ref{alg:streaming}). The first actions are therefore available after $|\mathcal{C}_1|$ tokens rather than after all $L$. After $s$ actions, the policy is queried with a new observation.

\begin{figure}[t]
  \centering
  \includegraphics[width=\columnwidth]{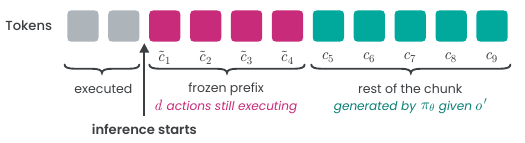}
  \caption{\textbf{Real-time chunking with action tokens.} Gray tokens encode actions already executed. When inference starts, the next $d$ actions will execute before the new chunk is ready, so NAT encodes these actions as the start of the new chunk and freezes the resulting first $P$ tokens ($\tilde{c}_{1:P}$). The policy generates only the remaining tokens $c_{P+1:L}$, given the new observation $o'$. Because the decoder is causal, the frozen prefix decodes to the actions being executed.}
  \label{fig:rtc}
\end{figure}

\textbf{Real-time chunking with tokens.} While the policy predicts the next chunk, the robot keeps executing the current one, so the next chunk must continue from the actions that run in the meantime. Because NAT's tokens are ordered in time and its decoder is causal, this can be done through the tokens alone, in three steps (Figure~\ref{fig:rtc}). (1)~Let $\tilde{a}_{1:d}$ be the last $d$ actions of the current chunk that will still execute, with $d$ a multiple of the $k$ actions per latent. NAT encodes them as the start of a new chunk, padded by repeating the last action, and their first $P$ tokens, $\tilde{c}_{1:P}$, form a fixed prefix.(2)~Given the new observation $o'$, the policy starts from this prefix and generates only the remaining tokens, $c_i \sim \pi_\theta(\cdot \mid o', c_{<i})$ for $i > P$. (3)~Because the decoder is causal, the first $d$ actions of the new chunk depend only on the prefix and approximately reproduce $\tilde{a}_{1:d}$, while the later actions are predicted to continue them. The robot therefore switches to the new chunk after $\tilde{a}_{1:d}$ without less of a jump. This gives the main benefits of real-time chunking (RTC)~\cite{black2026rtc}, which inpaints the executing actions into a flow policy's sampling, without any change to training or sampling, since conditioning on earlier tokens is what an autoregressive policy already does. In our NAT~VLAs, RTC lowers the mean absolute difference between the last executed action of a chunk and the first of the next (Table~\ref{tab:rtc}).

\begin{algorithm}[t]
\caption{Streaming NAT policy inference}
\label{alg:streaming}
\begin{algorithmic}[1]
\Require Observation $o$; policy $\pi_\theta$; NAT decoder $D_\psi$; execution horizon $s$
\State $c \gets [\,\texttt{BOS}\,]$
\For{$t = 1, \dots, T$}
    \For{each token position $i$ in $\mathcal{C}_t$}
        \State Sample $c_i \sim \pi_\theta(\cdot \mid o, c_{<i})$ and append it to $c$
    \EndFor
    \State Recover $\hat{z}_{1:t}$ from $c$ and decode the actions of latent $t$ with $D_\psi$
    \State Send these $k$ actions to the robot for execution
    \If{$tk \ge s$} \textbf{break} \Comment{the rest of the chunk will not be executed}
    \EndIf
\EndFor
\end{algorithmic}
\end{algorithm}

\section{Experiments}
\label{sec:experiments}

\subsection{Experimental Setup}

\begin{figure}[t]
  \centering
  \includegraphics[width=\columnwidth]{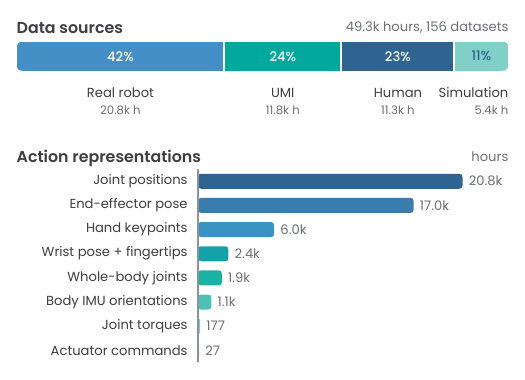}
  \caption{\textbf{Action corpus.} Hours by source (top) and action representation (bottom).}
  \label{fig:corpus-composition}
\end{figure}

\textbf{Action corpus.} The corpus has 49k hours (5.5B action frames) from 156 datasets: real-robot teleoperation, simulation, handheld grippers (UMI), and human hands (Figure~\ref{fig:corpus-composition}). Each dataset is normalized per dimension to $[-1, 1]$; chunks span $H = 32$ steps in groups of 7 dimensions (Appendix~\ref{app:corpus}).

\textbf{Zero-shot corpus.} The corpus excludes the evaluation robots (Franka Panda and Sawyer), so the tokenizers are not trained on them, except in Tables~\ref{tab:ablation-architecture} and~\ref{tab:ablation-sweeps}. Unless stated otherwise, tokenizers are trained on a fixed 1\% subset.

\textbf{Tokenizers.} We compare NAT with FAST~\cite{pertsch2025fast}, OAT~\cite{liu2026oat}, ActionCodec~\cite{dong2026actioncodec}, and NAC~\cite{jawaid2026nac}. All use the same data, a 51{,}200-token vocabulary, and 28 tokens per chunk (FAST averages 29.8 tokens); the neural tokenizers have 63M parameters.

\textbf{Policies.} Each frozen tokenizer drives the same autoregressive policy (Section~\ref{sec:policy}; 162M or 280M parameters, Table~\ref{tab:policy-hyperparameters}), which predicts 32-step chunks and executes 16. A diffusion policy~\cite{chi2025diffusion} with the same backbone serves as a non-autoregressive reference (Appendix~\ref{app:hyperparameters}).

\textbf{Benchmarks.} We evaluate on LIBERO-Long~\cite{liu2023libero}, Meta-World~\cite{yu2020metaworld}, RoboMimic~\cite{mandlekar2022matters}, and RoboCasa~\cite{nasiriany2024robocasa} with their standard protocols (Appendix~\ref{app:evaluation}), reporting mean success and 95\% confidence intervals over 4 seeds.

\subsection{How Does NAT Compare with Prior Tokenizers?}
\label{sec:exp-comparison}

\textbf{Prior tokenizers.} With one cross-embodiment policy per tokenizer (Table~\ref{tab:comparison}), NAT has the highest mean success (47.0\%), ahead of FAST (43.4\%), the other neural tokenizers (29.0--40.9\%), and the diffusion policy (35.9\%). NAT is best on LIBERO-Long and RoboCasa, FAST on Meta-World, and OAT on RoboMimic. OAT and NAC solve under 10\% of Meta-World, whose actions fill a small part of the normalized range and so magnify reconstruction error (Appendix~\ref{app:evaluation}). With a LIBERO-only policy (Table~\ref{tab:comparison-libero}), NAT reaches 77.9\% (FAST: 66.2\%).

\textbf{Tokens and generation rate.} On LIBERO, which runs at 20\,Hz, representing one second of action takes 9 tokens with the universal NAT, 24 with FAST+~\cite{pertsch2025fast} (2.8$\times$ more), and 140 with per-dimension binning (16$\times$ more).

\subsection{What Data Does an Action Tokenizer Need?}
\label{sec:exp-data}

\begin{figure}[t]
  \centering
  \includegraphics[width=\columnwidth]{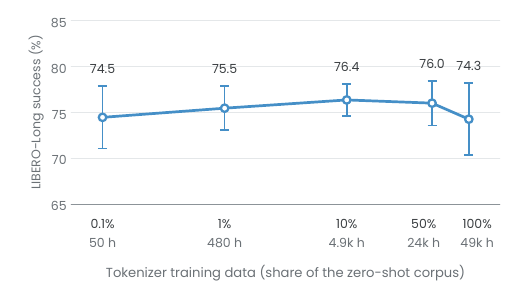}
  \caption{\textbf{NAT is data-efficient.} LIBERO-Long success of a fixed policy as NAT is trained on growing shares of the zero-shot corpus with the same number of updates. Error bars are 95\% confidence intervals.}
  \label{fig:data-scaling}
\end{figure}

These experiments train tokenizers on subsets of the zero-shot corpus and evaluate on LIBERO-Long~\cite{liu2023libero}, with a LIBERO-only policy for Tables~\ref{tab:comparison-libero} and~\ref{tab:data-mix} and a policy trained on the four LIBERO suites for Figure~\ref{fig:data-scaling} (Table~\ref{tab:policy-hyperparameters}).

\begin{table*}[t]
  \centering
  \begin{minipage}[t]{0.5\linewidth}
    \centering
    \caption{Tokenizers trained on all data or on human data only}
    \label{tab:comparison-libero}
    \footnotesize
    \setlength{\tabcolsep}{3pt}
    \begin{threeparttable}
    \begin{tabular*}{\linewidth}{@{\extracolsep{\fill}}lcccc}
      \toprule
       & \multicolumn{2}{c}{Recon.\ error} & \multicolumn{2}{c}{Success (\%)} \\
      \cmidrule(lr){2-3} \cmidrule(lr){4-5}
      Method & All & Human & All & Human \\
      \midrule
      NAT (ours) & 1.44 & 1.76 & $\mathbf{77.9 \pm 1.8}$ & $\mathbf{74.4 \pm 2.2}$ \\
      FAST \cite{pertsch2025fast} & 0.95 & 0.95 & $66.2 \pm 2.1$ & $67.0 \pm 3.5$ \\
      OAT \cite{liu2026oat} & 3.10 & 3.93 & $61.5 \pm 1.8$ & $38.6 \pm 1.3$ \\
      ActionCodec \cite{dong2026actioncodec} & 2.47 & 3.45 & $58.4 \pm 1.2$ & $40.9 \pm 2.8$ \\
      NAC \cite{jawaid2026nac} & 2.56 & 4.04 & $57.8 \pm 2.8$ & $25.1 \pm 1.9$ \\
      \bottomrule
    \end{tabular*}
    \tabnote{One LIBERO-only policy per tokenizer, evaluated on LIBERO-Long. \emph{All}: tokenizer trained on all source types; \emph{Human}: on human-hand data only. Recon.\ error: mean absolute error ($\times 10^{-2}$) on LIBERO-10 arm actions.}
    \end{threeparttable}
  \end{minipage}\hfill
  \begin{minipage}[t]{0.46\linewidth}
    \centering
    \caption{Data mixture ablation}
    \label{tab:data-mix}
    \footnotesize
    \begin{threeparttable}
    \begin{tabular*}{\linewidth}{@{\extracolsep{\fill}}lcc}
      \toprule
      Source type & Only this type & This type removed \\
      \midrule
      Real robot & $77.8 \pm 4.0$ & $75.9 \pm 3.2$ \\
      Sim robot & -- & $77.0 \pm 5.1$ \\
      UMI \cite{chi2024umi} & $76.0 \pm 2.4$ & $77.2 \pm 3.6$ \\
      Human hands & $74.4 \pm 2.2$ & $74.8 \pm 2.5$ \\
      UMI and human hands & -- & $73.8 \pm 2.7$ \\
      \bottomrule
    \end{tabular*}
    \tabnote{NAT trained on one source type only or with source types removed, in the setup of Table~\ref{tab:comparison-libero}; with all source types it reaches $77.9 \pm 1.8$\%. Dashes: not run.}
    \end{threeparttable}
  \end{minipage}
\end{table*}

\textbf{Human data alone.} Trained on human hands only (Table~\ref{tab:comparison-libero}, \emph{Human}), NAT keeps 74.4\% and FAST, whose BPE is not learned, is unchanged, while OAT, ActionCodec, and NAC lose 18--33 points. NAT's reconstruction error on LIBERO grows by 22\%, versus 27\% for OAT and 40\% for ActionCodec. A human-only NAT is also close to the all-data NAT with a cross-embodiment policy (Appendix~\ref{app:human-multi}).

\textbf{Data mix.} Every subset of source types stays within 4.1 points of the full mix and above FAST (Table~\ref{tab:data-mix}). Removing human-hand data causes the largest drop (3.1 points).

\textbf{Data amount.} With the policy fixed, training NAT on 0.1\% to 100\% of the corpus (50 to 49k hours, same number of updates) keeps success between 74.3\% and 76.4\% (Figure~\ref{fig:data-scaling}).

\subsection{Which Architectural Choices Matter?}
\label{sec:ablations}

\begin{table}[!t]
  \centering
  \caption{Architecture ablations on LIBERO-Long}
  \label{tab:ablation-architecture}
  \footnotesize
  \setlength{\tabcolsep}{3pt}
  \begin{threeparttable}
  \begin{tabular*}{\linewidth}{@{\extracolsep{\fill}}lcccc}
    \toprule
    Component & NAT & Success & Alternative & Success \\
    \midrule
    Quantizer$^{\dagger}$ & MS-FSQ & $69.3 \pm 2.2$ & MS-RVQ & $60.5 \pm 1.2$ \\
    Output head & Linear & $75.8 \pm 3.6$ & ISTFT & $75.6 \pm 0.5$ \\
    Causality & Causal & $75.8 \pm 3.6$ & Non-causal & $77.5 \pm 3.1$ \\
    \midrule
    Scales (tokens) & 1 (8) & 2 (12) & \textbf{4 (28)} & 8 (60) \\
    Success & $12.9 \pm 0.8$ & $56.3 \pm 4.3$ & $75.8 \pm 3.6$ & $74.9 \pm 1.6$ \\
    \bottomrule
  \end{tabular*}
  \tabnote{Success (\%) of NAT's choice and an alternative for each component. $^{\dagger}$Run on an earlier reference (Appendix~\ref{app:sweeps}).}
  \end{threeparttable}
\end{table}

Table~\ref{tab:ablation-architecture} ablates NAT's changes to NAC~\cite{jawaid2026nac} and the number of scales. Here and in Table~\ref{tab:ablation-sweeps} only, the 1\% subset comes from an earlier corpus that includes the evaluation robots.

\textbf{Scales and quantization.} Success rises from 12.9\% with one scale to 75.8\% with four, with no gain from eight. Replacing FSQ with NAC's RVQ costs 8.8 points. We suspect two causes, not isolated here: RVQ's learned codebooks are prone to collapse, leaving codes unused~\cite{mentzer2024fsq}; and FSQ's fixed grid makes codes that share most channel levels decode to similar latents, so a policy's confusions between them cost less, while RVQ's codes have no such structure.

\textbf{Output head and causality.} The ISTFT and linear heads perform similarly (75.6\% vs.\ 75.8\%), so we use the linear head. A non-causal NAT scores 1.7 points higher, within the 95\% intervals. Appendix~\ref{app:sweeps} gives further sweeps.

\subsection{Does NAT Help Vision-Language-Action Models?}
\label{sec:exp-vla}

\begin{figure}[!tb]
  \centering
  \includegraphics[width=\columnwidth]{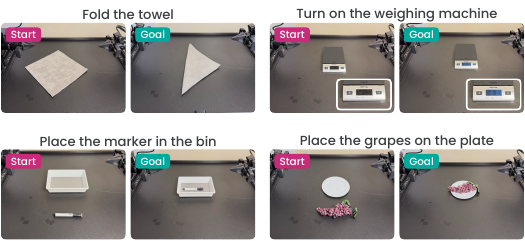}
  \caption{\textbf{Real-world tasks} on the Trossen AI Solo arm: start and goal states.}
  \label{fig:real-tasks}
\end{figure}

\textbf{Universal NAT.} Our VLAs use a \emph{universal NAT} with 4 latents (14 tokens) per chunk and 6{,}400 codes per scale, trained for one pass over the full zero-shot corpus (910M windows). Appendix~\ref{app:sweeps} varies the token and codebook budgets.

We pretrain two VLAs built on LFM2-VL-450M~\cite{liquidai2025lfm2}, one with our universal NAT (\emph{NAT~VLA}) and one with FAST+~\cite{pertsch2025fast} (\emph{FAST+~VLA}), otherwise identical. Both train on 4.8B tokens of MolmoAct2's seven-robot data mixture~\cite{fang2026molmoact2actionreasoningmodels} (1.8k hours), which includes DROID~\cite{droid}, BC-Z~\cite{bcz}, RT-1~\cite{brohan2023rt1}, and BridgeData~V2~\cite{bridgedata} (Appendix~\ref{app:vla}). On held-out data, NAT's action error under greedy decoding (temperature 0) is 16\% lower (Appendix~\ref{app:vla}). We fine-tune each jointly on four simulation benchmarks and on a real robot.

\begin{table*}[t]
  \centering
  \caption{One VLA fine-tuned on four benchmarks}
    \label{tab:vla-multi}
  \footnotesize
  \setlength{\tabcolsep}{4pt}
  \begin{threeparttable}
  \begin{tabular*}{\linewidth}{@{\extracolsep{\fill}}llccccc}
    \toprule
    Tokenizer & Acts via & LIBERO & Meta-World & RoboMimic & RoboCasa & Mean \\
    \midrule
    NAT (ours) & Tokens & $\mathbf{90.8 \pm 2.6}$ & $\mathbf{22.8 \pm 2.2}$ & $\mathbf{81.2 \pm 2.4}$ & $\mathbf{50.9 \pm 3.3}$ & $\mathbf{61.4}$ \\
    FAST+ & Tokens & $62.0 \pm 4.2$ & $20.9 \pm 2.4$ & $29.1 \pm 2.8$ & $37.6 \pm 3.2$ & $37.4$ \\
    NAT, no robot pretraining & Tokens & $86.0 \pm 2.8$ & $15.6 \pm 1.6$ & $72.5 \pm 2.6$ & $45.4 \pm 3.1$ & $54.9$ \\
    \midrule
    NAT (ours) & Flow expert & $\mathbf{89.2 \pm 2.8}$ & $\mathbf{45.7 \pm 2.5}$ & $\mathbf{89.8 \pm 2.0}$ & $\mathbf{69.5 \pm 3.0}$ & $\mathbf{73.6}$ \\
    FAST+ & Flow expert & $89.0 \pm 2.5$ & $43.9 \pm 2.5$ & $84.7 \pm 2.3$ & $67.9 \pm 3.0$ & $71.4$ \\
    \bottomrule
  \end{tabular*}
  \tabnote{Success (\%) at 50k steps with 95\% bootstrap intervals, acting through action tokens or a flow-matching expert. All rows use robot-pretrained backbones except \emph{no robot pretraining}, which is fine-tuned from the base LFM2-VL-450M with the same recipe. Bold: best within each group.}
  \end{threeparttable}
\end{table*}

\begin{table*}[t]
  \centering
  \caption{Action latency (ms) on an RTX 4090}
    \label{tab:latency}
  \footnotesize
  \setlength{\tabcolsep}{3pt}
  \begin{threeparttable}
  \begin{tabular*}{\linewidth}{@{\extracolsep{\fill}}llccccc}
    \toprule
    & & & & \multicolumn{2}{c}{Action tokens} & Flow head \\
    \cmidrule(lr){5-6}\cmidrule(lr){7-7}
    Model & Tokenizer & Size & Per token & First actions & Full chunk & Full chunk \\
    \midrule
    NAT~VLA (ours) & NAT & 0.45B & $2.0 \pm 0.0$ & $\mathbf{19.0 \pm 0.5}^{\ast}$ & $39.5 \pm 0.6^{\ast}$ & $42.0 \pm 0.9$ \\
    FAST+~VLA & FAST+ & 0.45B & $2.2 \pm 0.0$ & $97.5 \pm 37.6$ & $97.5 \pm 37.6$ & $\mathbf{46.5 \pm 12.6}$ \\
    \midrule
    GR00T N1.7~\cite{nvidia2025groot} & NAT$^{\dagger}$ & 3B & $4.1 \pm 0.0$ & $\mathbf{44.7 \pm 1.0}$ & $85.2 \pm 1.0$ & $56.6 \pm 1.2$ \\
    $\pi_{0.5}$~\cite{black2025pi0.5} & NAT$^{\dagger}$ & 3B & $6.5 \pm 0.0$ & $\mathbf{69.0 \pm 1.7}$ & $134.5 \pm 1.7$ & $72.6 \pm 0.2$ \\
    MolmoAct2~\cite{fang2026molmoact2actionreasoningmodels} & NAT$^{\dagger}$ & 5B & $12.6 \pm 0.0$ & $\mathbf{100.8 \pm 0.3}$ & $226.4 \pm 0.4$ & $132.4 \pm 1.3$ \\
    \bottomrule
  \end{tabular*}
  \tabnote{Batch size 1, CUDA graphs; mean $\pm$ s.d.\ over 50 frames for our models and 30 runs for the others. \emph{First actions}: 8 actions from 4 NAT tokens; \emph{full chunk}: 14 NAT tokens. Bold: faster route to the first actions. $^{\ast}$Includes detokenization. $^{\dagger}$Latency analysis only: these models are not trained with NAT; we time 4 and 14 decoding steps.}
  \end{threeparttable}
\end{table*}

\begin{table}[t]
  \centering
  \caption{VLA success on a real Trossen AI Solo arm}
  \label{tab:real-world}
  \footnotesize
  \setlength{\tabcolsep}{4pt}
  \begin{threeparttable}
  \begin{tabular*}{\linewidth}{@{\extracolsep{\fill}}lccccc}
    \toprule
    & Fold & Turn on & Marker & Grapes & \\
    & towel & scale & in bin & on plate & Mean \\
    \midrule
    NAT (ours) & \textbf{90} & \textbf{70} & \textbf{40} & 20 & \textbf{55} \\
    FAST+ & 50 & 20 & 0 & \textbf{70} & 35 \\
    \bottomrule
  \end{tabular*}
  \tabnote{Success (\%) over 10 trials per task. The tasks are shown in Figure~\ref{fig:real-tasks}.}
  \end{threeparttable}
\end{table}

\textbf{Multiple benchmarks.} We fine-tune each backbone jointly on the four benchmarks of Table~\ref{tab:comparison}, using all four LIBERO suites. Acting through tokens with the same pretraining and fine-tuning, NAT leads on all four and raises mean success from 37.4\% to 61.4\% (Table~\ref{tab:vla-multi}); on LIBERO it reaches 90.8\% against 62.0\% for FAST+, while on Meta-World the intervals overlap. Without robot pretraining, the NAT VLA reaches 54.9\%, still above the pretrained FAST+ VLA; robot pretraining through NAT tokens adds 6.5 points (4.8 on LIBERO), with non-overlapping intervals on Meta-World and RoboMimic.

\textbf{Real robot.} We fine-tune both VLAs on 481 teleoperated demonstrations on a Trossen AI Solo arm. The universal NAT's corpus includes our own Trossen AI recordings. The four tasks are folding a towel, turning on a scale, putting a marker in a bin, and placing grapes on a plate (Figure~\ref{fig:real-tasks}). Each policy runs 10 trials per task from the demonstrations' median start pose, each scored by an operator (Appendix~\ref{app:vla}). NAT succeeds in 55\% of trials and FAST+ in 35\% (Table~\ref{tab:real-world}). NAT leads on three of the four tasks. FAST+ outperforms NAT on placing grapes (70\% vs.\ 20\%), although NAT's reconstruction error on the grape demonstrations is small (MSE 0.0005) and its policy's greedily sampled actions have lower reconstruction error than FAST+'s (MSE 0.021 vs.\ 0.038). In failed trials NAT~VLA did not appear to lower the gripper far enough and closed above the grapes; because each chunk's 16 actions execute open loop, such an offset may go uncorrected.

\subsection{How Much Does a Flow Action Head Help?}
\label{sec:exp-flow}

Many VLAs add a flow-matching action head with the aim of reducing latency and discretization error, so we evaluate the performance of one on top of our VLA. Following the MolmoAct2 recipe~\cite{fang2026molmoact2actionreasoningmodels}, we add a flow-matching action head~\cite{lipman2023flow} trained with knowledge insulation~\cite{driess2026knowledge}: the head attends to the language model's activations, but its gradients do not reach the language model, so the backbone keeps training on action tokens. Each pretrained model is first mid-trained with the head and then fine-tuned with tokens and head together (Appendix~\ref{app:vla}).

\textbf{Latency.} FAST+ uses 6 to 73 tokens per chunk (38.8 on average) and decodes only after the whole chunk. NAT uses 14, and its first 4 decode to 8 actions, which take 267\,ms to execute at 30\,Hz; the rest of the chunk, and the next one (Section~\ref{sec:policy}), are generated while the robot moves. Table~\ref{tab:latency} times this on NAT~VLA (50 samples). It also times the same decoding steps on the language models of larger VLAs, as a latency analysis only. NAT~VLA has its first actions after 19.0\,ms, 5.1$\times$ sooner than FAST+~VLA and 2.2$\times$ sooner than its own flow expert. On these 3--5B models, generating NAT's first 4 tokens would take 45--101\,ms, comparable to or faster than their action heads, and a full chunk 85--226\,ms, under the 267\,ms the first actions take to execute.

\textbf{Success and cost.} We fine-tune each backbone with the flow head on the four benchmarks of Table~\ref{tab:vla-multi}. These runs add the mid-training stage and use MolmoAct2's post-training learning rates, so they differ from the token runs in more than the action head. Through the expert, NAT~VLA reaches 73.6\% mean success and FAST+~VLA 71.4\% (Table~\ref{tab:vla-multi}); the intervals overlap except on RoboMimic. The expert raises NAT's mean by 12.2 points, mostly on Meta-World and RoboCasa, and FAST+'s by 34.0. On LIBERO, NAT tokens and the expert are within the 95\% intervals (90.8\% vs.\ 89.2\%). The expert adds 77.7M parameters, a mid-training stage, and a denoising process at inference, while NAT's first actions arrive sooner than a chunk from our flow head (19.0 vs.\ 42.0\,ms) and, in our latency analysis, than the heads of GR00T N1.7 and MolmoAct2 (Table~\ref{tab:latency}), and its tokens keep the streaming and real-time chunking of Section~\ref{sec:policy}.

\section{Discussion}

Our results support four takeaways. First, a single neural action tokenizer can serve robots it has never seen, outperforming FAST and prior neural tokenizers without training on the evaluation robots. Second, an action tokenizer may not need robot data: human hand motion alone yields a tokenizer close to one trained on all sources, so large-scale human data could be used to train action tokenizers. Third, NAT tokens are an effective action representation for vision-language-action models, both for robot pretraining and for acting, where they outperform FAST+. Fourth, action tokens can be faster than flow action heads, and on some domains acting through tokens alone closes the performance gap to a flow head, though a gap remains on others.

\textbf{Limitations and future work.} Acting through NAT tokens still trails a flow-matching head on some benchmarks; further tuning and design choices, such as the token budget, decoding, and fine-tuning recipe, may close this gap fully. NAT encodes every chunk in a fixed number of tokens; a future study could set the number of tokens per chunk dynamically. NAT is also trained on a single chunk length, so tokenizing chunks of varying length remains to be explored. Our data-scaling experiment scales the tokenizer's training data while keeping the policy's LIBERO data fixed; a future study could explore the benefits of scaling the tokenizer corpus together with the policy's training corpus.

%% file: content/appendix.tex
\section{Action Corpus}
\label{app:corpus}

The corpus has 156 datasets (49.3k hours, 5.5B action frames) from four source types (Figure~\ref{fig:corpus-composition}). Each dataset is normalized per dimension to $[-1, 1]$ with min--max statistics from its training recordings, and its actions are split into groups of at most 7 dimensions (Section~\ref{sec:tokenization}). Padded dimensions and frames past the end of an episode are excluded from the loss. Dataset $k$ is sampled with probability proportional to $N_k^{\tau}$, where $N_k$ is its number of frames and $\tau$ a temperature. Validation and test sets (0.1\% each, from held-out recordings) are excluded from training, and no dataset contains the evaluation robots (Franka Panda and Sawyer).
Table~\ref{tab:corpus-summary} lists the collections in each source type. All collections were public when accessed, except our own Trossen AI recordings (5.3 hours) and two proprietary sets of human hand recordings: 1{,}956 hours from MicroAGI and 72 hours from DeepReach AI, both 21-joint hand poses from head-mounted capture.

\begin{table*}[t]
  \centering
  \caption{Action corpus by source type}
    \label{tab:corpus-summary}
  \footnotesize
  \begin{threeparttable}
  \begin{tabular}{p{0.15\linewidth}rrp{0.6\linewidth}}
    \toprule
    Source type & Datasets & Hours & Collections \\
    \midrule
    Real robot & 86 & 20{,}821 &
      Open X-Embodiment~\cite{o2024open}, BridgeData~V2~\cite{bridgedata}, BC-Z~\cite{bcz}, QT-Opt~\cite{qtopt}, Language-Table~\cite{languagetable}, RH20T~\cite{rh20t}, AgiBot World~\cite{agibotworld}, RoboMIND~\cite{robomind}, RoboMIND~2.0~\cite{robomind2}, RoboCOIN~\cite{robocoin}, OpenNeo~\cite{openneo}, ABC-130k~\cite{abc130k}, AIRoA MoMa~\cite{airoamoma}, Galaxea Open-World~\cite{galaxea}, MolmoAct2~\cite{fang2026molmoact2actionreasoningmodels}, WEB~\cite{webdataset}, OpenArms~\cite{openarms}, HIW-500~\cite{hiw500}, GR00T teleoperation~\cite{nvidia2025groot}, ActionNet~\cite{actionnet}, Humanoid Everyday~\cite{humanoideveryday}, Unitree G1/Z1~\cite{unitreedatasets}, ETRI G1~\cite{etrig1}, LeRobot community~\cite{cadene2026lerobot}, Open-H~\cite{openh}, TriFinger RRC~\cite{trifinger}, DexWild~\cite{dexwild}, HRDexDB~\cite{hrdexdb}, TDex~\cite{tdex}/TAVI~\cite{tavi}, TWIST2~\cite{twist2}, ReFlex~\cite{reflex}, EgoVerse~\cite{egoverse}, EgoSteer~\cite{egosteer}, PH$^2$D~\cite{qiu2025humanoid}, Trossen AI (ours)$^{\ast}$ \\
    \midrule
    Simulation & 18 & 5{,}408 &
      InternData~\cite{interndata}, BEHAVIOR-1K~\cite{behavior1k}, GraspXL~\cite{graspxl}, GR00T simulation~\cite{nvidia2025groot}, HOP~\cite{hop}, COMET~\cite{comet}, GenieSim~3~\cite{geniesim}, AgiBot World 2026~\cite{agibotworld2026}, DexTrack~\cite{dextrack}, LEAP-Lift~\cite{leaplift}, TriFinger RRC~\cite{trifinger}, Open-H~\cite{openh}, PH$^2$D~\cite{qiu2025humanoid} \\
    \midrule
    UMI & 7 & 11{,}755 &
      GenRobot~\cite{genrobot}, OpenNeo~\cite{openneo}, HY-Embodied~\cite{hyembodied}, HiFi-UMI~\cite{hifiumi}, FastUMI-100K~\cite{fastumi} \\
    \midrule
    Human & 45 & 11{,}347 &
      EgoVerse~\cite{egoverse}, EgoSteer~\cite{egosteer}, OpenEgo~\cite{openego}, VITRA~\cite{vitra}, MicroAGI$^{\ast}$, DeepReach AI$^{\ast}$, EgoDemo~\cite{egodemo}, Assembly101~\cite{assembly101}, PH$^2$D~\cite{qiu2025humanoid}, OMG~\cite{omg}, Chingmu MoCap~\cite{chingmu}, Tracker POV~\cite{trackerpov} \\
    \midrule
    Total & 156 & 49{,}331 & \\
    \bottomrule
  \end{tabular}
  \tabnote{Dataset counts treat each embodiment or recording variant of a collection separately. $^{\ast}$Proprietary.}
  \end{threeparttable}
\end{table*}

\section{Hyperparameters}
\label{app:hyperparameters}

Tables~\ref{tab:hyperparameters}--\ref{tab:policy-hyperparameters} list the tokenizer architecture, the tokenizer training runs, and the policies.

\begin{table}[H]
  \centering
  \caption{NAT architecture}
    \label{tab:hyperparameters}
  \footnotesize
  \begin{threeparttable}
  \setlength{\tabcolsep}{4pt}
  \begin{tabular}{lcc}
    \toprule
     & Standard & Universal \\
    \cmidrule(lr){2-2}\cmidrule(lr){3-3}
    \multicolumn{3}{l}{\textit{Tokenization}} \\
    Chunk & \multicolumn{2}{c}{$H = 32$ steps $\times$ 7 dimensions} \\
    Latents per chunk $T$ & 8 (4 steps each) & 4 (8 steps each) \\
    Latent dimension & \multicolumn{2}{c}{512} \\
    Quantizer & \multicolumn{2}{c}{Multi-scale FSQ, $M = 4$ scales} \\
    Scale spans & \multicolumn{2}{c}{$[2, 1, 1, 1]$ latents} \\
    FSQ codes per scale & 12{,}800 & 6{,}400 \\
    Tokens per chunk & 28 & 14 \\
    \midrule
    \multicolumn{3}{l}{\textit{Encoder (0.9M parameters)}} \\
    Downsampling, $L = 2$ & strides 4, 7 & strides 8, 7 \\
    Recurrent layer & \multicolumn{2}{c}{2-layer causal LSTM} \\
    \midrule
    \multicolumn{3}{l}{\textit{Decoder (62M parameters)}} \\
    Width & \multicolumn{2}{c}{768} \\
    ResBlocks around attention & \multicolumn{2}{c}{$R = 2$ before and 2 after} \\
    Causal attention window & \multicolumn{2}{c}{8 latents} \\
    ConvNeXt blocks & \multicolumn{2}{c}{$N = 12$} \\
    Output values per latent & 28 & 56 \\
    \bottomrule
  \end{tabular}
  \tabnote{Standard: the NAT of Tables~\ref{tab:comparison}--\ref{tab:data-mix}. Universal: the NAT of the VLAs (Section~\ref{sec:exp-vla}). Codes per scale for the other runs are in Table~\ref{tab:tokenizer-hyperparameters}. $L$, $R$, and $N$ correspond to Figure~\ref{fig:architecture}.}
  \end{threeparttable}
\end{table}

\begin{table}[H]
  \centering
  \caption{Tokenizer training}
    \label{tab:tokenizer-hyperparameters}
  \footnotesize
  \setlength{\tabcolsep}{3pt}
  \begin{threeparttable}
  \begin{tabular}{lccc}
    \toprule
                         & Standard & Data scaling & Ablations \\
    \midrule
    Used in & Tables~\ref{tab:comparison}--\ref{tab:data-mix} & Figure~\ref{fig:data-scaling} & Tables~\ref{tab:ablation-architecture}, \ref{tab:ablation-sweeps} \\
    Data & 1\% of recordings & 0.1--100\% & 1\% of recordings$^{\ast}$ \\
    Sampling temp. & $\tau = 0.5$ & $\tau = 1$ & $\tau = 0.5$ \\
    FSQ codes per scale & 12{,}800 & 512 & 512 or 2{,}048 \\
    Batch size & 1{,}024 & 1{,}024 & 256 \\
    Steps & 30{,}000 & 165{,}947 & 30{,}000 \\
    Warm-up steps & 9{,}000 & 49{,}784 & 9{,}000 \\
    \bottomrule
  \end{tabular}
  \tabnote{$^{\ast}$Of a larger corpus that also includes the evaluation robots.}
  \end{threeparttable}
\end{table}

\begin{table*}[t]
  \centering
  \caption{Policies}
    \label{tab:policy-hyperparameters}
  \footnotesize
  \setlength{\tabcolsep}{4pt}
  \begin{threeparttable}
  \begin{tabular*}{\linewidth}{@{\extracolsep{\fill}}lccc}
    \toprule
                     & LIBERO-only & Cross-embodiment & LIBERO suites \\
    \midrule
    Used in & Tables~\ref{tab:comparison-libero}, \ref{tab:data-mix} & Table~\ref{tab:comparison} & Fig.~\ref{fig:data-scaling}, Tables~\ref{tab:ablation-architecture}, \ref{tab:ablation-sweeps} \\
    Training data & LIBERO-10 & Four benchmarks & Four LIBERO suites \\
    Parameters & 280M & 162M & 162M \\
    Width / layers / heads & 1{,}024 / 16 / 16 & 768 / 16 / 12 & 768 / 16 / 12 \\
    Steps & 25{,}000 & 50{,}000 & 25{,}000 \\
    Batch size & 256 & 256 & 256 \\
    \midrule
    \multicolumn{4}{l}{\textit{Shared by all policies}} \\
    Observations & \multicolumn{3}{l}{Last 2 steps; ResNet-18 with spatial softmax per camera,} \\
                     & \multicolumn{3}{l}{linear proprioception, frozen BERT instruction} \\
    Execution & \multicolumn{3}{l}{32-step chunk decoded greedily, first 16 steps executed} \\
    Vocabulary & \multicolumn{3}{l}{51{,}200 tokens for compared tokenizers; else $M\times$ codes (Table~\ref{tab:tokenizer-hyperparameters})} \\
    \bottomrule
  \end{tabular*}
  \tabnote{The human-only cross-embodiment policy of Table~\ref{tab:human-multi} uses the LIBERO-only architecture (280M) for 25{,}000 steps.}
  \end{threeparttable}
\end{table*}

\textbf{Baselines.} OAT, ActionCodec, and NAC keep their original configurations but are matched to NAT's data, 28 tokens per chunk, 51{,}200-token vocabulary, and 62--63M parameters. FAST's 51{,}200-token BPE vocabulary is fitted on the same corpus.

\section{Evaluation Protocols}
\label{app:evaluation}

All benchmarks follow their reference protocols with 4 evaluation seeds (0--3) and 50 trials per task per seed. \textbf{LIBERO-Long}: 10 tasks with the official initial states. \textbf{Meta-World}: box-close, coffee-pull, disassemble, and stick-pull (800 scripted demonstrations). \textbf{RoboMimic}: Lift, Can, Square, and ToolHang (800 proficient-human demonstrations). \textbf{RoboCasa}: CloseDrawer, TurnOffSinkFaucet, CoffeePressButton, and TurnOffMicrowave (216 human demonstrations). \emph{Mean} averages the four benchmarks within each seed and reports the mean and 95\% confidence interval over seeds.

Meta-World's actions are normalized by their training limits, which squeezes the executed commands into about a fifth of the tokenizer's range and magnifies reconstruction error. The reconstruction error in Table~\ref{tab:comparison-libero} is the round-trip mean absolute error on LIBERO-10 chunks, averaged over position and rotation dimensions in normalized units.

\begin{table*}[t]
  \centering
  \caption{Pretrained VLA backbones on held-out robot data}
    \label{tab:backbone}
  \footnotesize
  \setlength{\tabcolsep}{4pt}
  \begin{threeparttable}
  \begin{tabular*}{\linewidth}{@{\extracolsep{\fill}}lcccccccc}
    \toprule
    & & \multicolumn{3}{c}{NAT (ours)} & \multicolumn{3}{c}{FAST+} \\
    \cmidrule(lr){3-5}\cmidrule(lr){6-8}
    Robot & Steps & PPL & Greedy & Sampled & PPL & Greedy & Sampled \\
    \midrule
    YAM (bimanual) & 30 & 27.8 & \textbf{0.0250} & 0.0985 & 7.5 & 0.0314 & \textbf{0.0419} \\
    SO-100/101 & 30 & 70.9 & \textbf{0.0067} & 0.0307 & 11.1 & 0.0078 & \textbf{0.0071} \\
    MolmoAct Franka & 10 & 91.5 & \textbf{0.0551} & \textbf{0.1163} & 10.7 & 0.0784 & 0.1236 \\
    DROID & 15 & 11.3 & \textbf{0.0479} & \textbf{0.1000} & 6.9 & 0.0579 & 0.1087 \\
    BC-Z & 10 & 274.8 & 0.1280 & 0.2222 & 25.7 & \textbf{0.1216} & \textbf{0.1277} \\
    RT-1 & 3 & 94.9 & \textbf{0.1009} & 0.2088 & 8.1 & 0.1284 & \textbf{0.1107} \\
    Bridge & 5 & 28.6 & 0.1075 & \textbf{0.1702} & 3.5 & \textbf{0.1068} & 0.2357 \\
    \midrule
    Mix-weighted & & 31.7 & \textbf{0.0337} & 0.0867 & 8.4 & 0.0400 & \textbf{0.0622} \\
    \bottomrule
  \end{tabular*}
  \tabnote{\emph{Greedy}/\emph{Sampled}: action MSE under greedy and temperature-1 decoding over MolmoAct2's horizon (\emph{Steps}); lower is better.}
  \end{threeparttable}
\end{table*}

\section{Design Sweeps}
\label{app:sweeps}

\begin{table}[!t]
  \centering
  \caption{NAT design sweeps on LIBERO-Long}
    \label{tab:ablation-sweeps}
  \footnotesize
  \begin{threeparttable}
  \begin{tabular}{llc}
    \toprule
    Sweep & Setting & Success (\%) \\
    \midrule
    Codes per scale & 512 & $76.6 \pm 1.4$ \\
                             & 1{,}024 & $75.2 \pm 2.3$ \\
                             & 2{,}048 & $75.8 \pm 3.6$ \\
                             & 4{,}096 & $76.5 \pm 2.8$ \\
                             & 8{,}192 & $75.0 \pm 4.2$ \\
    \midrule
    Scale layout & $[2,1,1,1]$, 28 tokens & $76.6 \pm 1.4$ \\
    (512 codes) & $[2,1,1]$, 20 tokens & $68.2 \pm 3.8$ \\
                             & $[2,1,1,1]$, 4 latents, 14 tokens & $73.3 \pm 0.9$ \\
                             & Pyramid $[4,2,1]$, 14 tokens & $65.1 \pm 3.0$ \\
                             & Pyramid $[8,4,2,1]$, 15 tokens & $61.1 \pm 3.0$ \\
                             & Pyramid $[16,\dots,1]$, 31 tokens & $70.2 \pm 1.8$ \\
                             & Pyramid $[32,\dots,1]$, 63 tokens & $67.6 \pm 3.9$ \\
    \midrule
    Policy size & 162M & $75.8 \pm 3.6$ \\
                             & 21M & $67.1 \pm 3.4$ \\
                             & 40M & $72.8 \pm 2.0$ \\
                             & 81M & $75.5 \pm 1.8$ \\
    \midrule
    Tokenizer size & 63M & $75.8 \pm 3.6$ \\
                             & 5M & $69.8 \pm 2.5$ \\
                             & 124M & $75.8 \pm 2.5$ \\
                             & 243M & $77.2 \pm 2.6$ \\
    \midrule
    Action frequency & Native rates & $76.6 \pm 1.4$ \\
    (512 codes) & Resampled to 20\,Hz & $74.0 \pm 2.3$ \\
                             & Resampled to 30\,Hz & $70.6 \pm 1.7$ \\
    \bottomrule
  \end{tabular}
  \tabnote{Each block varies one setting; scales are spans over a chunk's latents.}
  \end{threeparttable}
\end{table}

\textbf{Token budget and codebook size.} Table~\ref{tab:token-budget} varies the number of latents per chunk, which sets the number of tokens, and the number of codes per scale. Halving the tokens from 28 to 14 does not reduce success here (74.4\% vs.\ 75.8\%), but it does in the 512-code sweep of Table~\ref{tab:ablation-sweeps} (76.6\% vs.\ 73.3\%), so the effect may depend on the codebook size or data. The 14-token NAT reaches 78.0\% when trained on the full corpus (the universal NAT also uses 14 tokens, but 6{,}400 codes and a longer training run). Halving again to 7 tokens drops success to 55.1\%, as LIBERO arm reconstruction error rises from 0.025 to 0.040. Codebook size has little effect from 512 to 12{,}800 codes in our runs (Table~\ref{tab:ablation-sweeps}; 74.2\% with 6{,}400 and 75.8\% with 12{,}800 here), but 256 codes reduce success to 64.5\%.

\begin{table}[!t]
  \centering
  \caption{Token budget and codebook size on LIBERO-Long}
    \label{tab:token-budget}
  \footnotesize
  \setlength{\tabcolsep}{4pt}
  \begin{threeparttable}
  \begin{tabular}{cccc}
    \toprule
    Tokens & Latents & Codes & Success (\%) \\
    \midrule
    28 & 8 & 12{,}800 & 74.4 \\
    14 & 4 & 12{,}800 & \textbf{75.8} \\
    14 & 4 & 6{,}400 & 74.2 \\
    7 & 2 & 12{,}800 & 55.1 \\
    28 & 8 & 256 & 64.5 \\
    \bottomrule
  \end{tabular}
  \tabnote{NAT trained on 1\% of the zero-shot corpus. Tokens per $32 \times 7$ chunk; codes per scale.}
  \end{threeparttable}
\end{table}

These ablations train each tokenizer on the zero-shot corpus for 30{,}000 steps of 1{,}024 windows, then one 162M cross-embodiment policy, evaluated on LIBERO-Long over 4 seeds.

\textbf{Universal NAT.} The universal NAT trains for 222{,}289 steps of 4{,}096 windows, with learning rate $2\times10^{-4}$ and cosine decay.

\textbf{Architecture ablations.} Tables~\ref{tab:ablation-architecture} and~\ref{tab:ablation-sweeps} train tokenizers on 1\% of an earlier corpus that includes the evaluation robots (30{,}000 steps, batch 256), then a 162M policy on the four LIBERO suites (25{,}000 steps), evaluated on LIBERO-Long over 4 seeds. Each block changes one setting of the reference NAT, which used 2{,}048 codes per scale (512 for the scale-layout and frequency sweeps).

The quantizer comparison used an earlier reference that delays the coarsest token until both latents it covers are seen, so that no token depends on future actions. Without this delay, the coarsest token can depend on up to 4 future steps when tokenizing a ground-truth chunk, but not when decoding generated tokens. The RVQ variant uses NAC's commitment loss.

\section{Human-Only NAT on Multiple Benchmarks}
\label{app:human-multi}

Table~\ref{tab:human-multi} repeats the multi-benchmark comparison with a NAT trained only on human-hand data. Both NATs drive the same cross-embodiment policy, which uses the LIBERO-only architecture for 25{,}000 steps (Table~\ref{tab:policy-hyperparameters}), so success is lower than in Table~\ref{tab:comparison}. The human-only NAT is within 0.8 points of the all-data NAT on every benchmark (mean 38.0\% vs.\ 38.5\%).

\begin{table}[H]
  \centering
  \caption{NAT trained on human-hand data only}
    \label{tab:human-multi}
  \footnotesize
  \begin{threeparttable}
  \begin{tabular*}{\linewidth}{@{\extracolsep{\fill}}lcc}
    \toprule
    Benchmark & All source types & Human hands only \\
    \midrule
    LIBERO-Long & $53.0 \pm 4.1$ & $52.9 \pm 2.0$ \\
    Meta-World & $32.4 \pm 1.4$ & $31.8 \pm 2.9$ \\
    RoboMimic & $34.0 \pm 2.5$ & $33.2 \pm 2.6$ \\
    RoboCasa & $34.6 \pm 3.5$ & $34.0 \pm 6.2$ \\
    \midrule
    Mean & $38.5 \pm 1.9$ & $38.0 \pm 1.8$ \\
    \bottomrule
  \end{tabular*}
  \tabnote{Success (\%) in the multi-benchmark setting of Table~\ref{tab:comparison}.}
  \end{threeparttable}
\end{table}

\section{Vision-Language-Action Models}
\label{app:vla}

\textbf{Pretraining.} NAT~VLA and FAST+~VLA are LFM2-VL-450M models pretrained with MolmoAct2's trainer and data pipeline~\cite{fang2026molmoact2actionreasoningmodels} on its seven-robot mixture (Table~\ref{tab:vla-mixture}): 1{,}843 hours in 333{,}729 episodes, of which 16{,}184 are held out. The two runs differ only in the tokenizer. Both train for 95{,}315 steps on batches of 12 packed 4{,}200-token sequences (4.8B tokens), with learning rates $5\times10^{-5}$ (language model) and $2.5\times10^{-5}$ (vision encoder and connector). Actions are 32-step chunks at each dataset's native rate, normalized to $[-1,1]$, and both tokenizers share 25{,}600 added action-token embeddings. NAT~VLA uses the universal NAT (14 tokens per group of up to 7 dimensions); FAST+~VLA uses the released FAST+ tokenizer on the whole chunk. Both are trained with MolmoAct2's language-model loss rather than Eq.~(\ref{eq:policy}): the loss covers each sample's answer, an action-start token, the action tokens, an action-end token, and an end-of-sequence token; each sample's token losses are weighted by $1/\sqrt{N}$ for its $N$ answer tokens, with the weights normalized over the batch; and a z-loss~\cite{chowdhery2023palm} with coefficient $10^{-4}$ is added.

\begin{table}[!t]
  \centering
  \caption{VLA pretraining mixture}
    \label{tab:vla-mixture}
  \footnotesize
  \setlength{\tabcolsep}{4pt}
  \begin{threeparttable}
  \begin{tabular}{lrrrr}
    \toprule
    Robot & Rate & Episodes & Held out & Hours \\
    \midrule
    YAM (bimanual)~\cite{fang2026molmoact2actionreasoningmodels} & 30\% & 32{,}246 & 2{,}386 & 704.1 \\
    SO-100/101~\cite{cadene2026lerobot} & 30\% & 36{,}637 & 1{,}085 & 177.1 \\
    MolmoAct Franka~\cite{lee2025molmoact} & 2.5\% & 10{,}488 & 11 & 25.2 \\
    DROID~\cite{droid} & 30\% & 74{,}604 & 2{,}651 & 328.9 \\
    BC-Z~\cite{bcz} & 2.5\% & 39{,}350 & 2{,}938 & 152.0 \\
    RT-1~\cite{brohan2023rt1} & 2.5\% & 87{,}212 & 3{,}985 & 350.6 \\
    Bridge~\cite{bridgedata} & 2.5\% & 53{,}192 & 3{,}128 & 105.2 \\
    \midrule
    Total & & 333{,}729 & 16{,}184 & 1{,}843.0 \\
    \bottomrule
  \end{tabular}
  \tabnote{Rate: sampling rate in the mixture. Episodes and hours include the held-out episodes.}
  \end{threeparttable}
\end{table}

\textbf{Held-out evaluation.} Both models are evaluated at their final checkpoint on the same 128 held-out frames per robot (Table~\ref{tab:backbone}). We report perplexity per action token and the action mean squared error under greedy and temperature-1 decoding, in normalized units over MolmoAct2's horizon. The mix-weighted error weights robots by their sampling rates.

With greedy decoding, NAT has lower action error on five of seven robots and in mix-weighted error. With temperature-1 sampling, FAST+ is lower on four robots and in mix-weighted error, so sampling raises NAT's error more than FAST+'s.

\textbf{Real-world evaluation.} Both models are fine-tuned for 10{,}000 steps (batch 64) on 481 teleoperated demonstrations on a Trossen AI Solo arm, recorded at 30\,Hz with top and wrist cameras and absolute joint targets for six joints and a gripper. Both decode greedily, predict 32-step chunks, and execute 16 at 30\,Hz. The next chunk is predicted while the last 6 actions of the current one run. For safety, a chunk with a jump above 0.3\,rad between consecutive actions is not executed. Each policy ran 10 trials per task, each up to 33\,s from the demonstrations' median start pose and scored by an operator.

\textbf{Latency.} Table~\ref{tab:latency} times every model on one RTX 4090 in bfloat16 at batch size 1. Our models are timed on 50 distinct LIBERO frames; each baseline uses its released checkpoint and example input, timed 30 times after 3 warm-up runs. At batch size 1, eager PyTorch is dominated by kernel launches, so every model's decoding and action head are replayed as CUDA graphs, which produce the same outputs as eager execution. The baselines are GR00T N1.7 (its 16-layer language model, with Qwen3-VL-2B in place of the gated backbone of the same architecture), $\pi_{0.5}$ (\texttt{pi05\_libero}), and MolmoAct2 (\texttt{MolmoAct2-LIBERO}). Prompts are 237 tokens for our model, 469 for GR00T N1.7, 497 for MolmoAct2, and 968 for $\pi_{0.5}$. Baseline token timings exclude NAT's detokenization (about 3\,ms). FAST+~VLA uses 6 to 73 tokens per chunk, so its latency varies (median 85.3\,ms, 90th percentile 146.8\,ms).

\textbf{Multiple environments.} Each pretrained model is fine-tuned on action tokens only for 50{,}000 steps at batch 64, with learning rates $5\times10^{-5}$ (language model) and $2.5\times10^{-5}$ (vision encoder and connector), on an equal mixture of LIBERO (MolmoAct2's four-suite dataset), Meta-World, RoboMimic, and RoboCasa demonstrations. Policies predict 32-step chunks, decode greedily, and each simulator executes the first 10 actions of every chunk. All results are at step 50{,}000. LIBERO uses 400 episodes (10 tasks per suite, 10 initial states each), RoboMimic and RoboCasa 800 each, and Meta-World 1{,}000 (5 seeds), with 95\% bootstrap intervals. The \emph{no robot pretraining} row uses the same recipe from the released LFM2-VL-450M, with new action-token embeddings.

\emph{Flow expert.} For the flow-expert rows of Table~\ref{tab:vla-multi}, each pretrained model is first mid-trained for 23{,}829 steps with MolmoAct2's knowledge-insulation recipe~\cite{driess2026knowledge}, which adds a 77.7M-parameter flow-matching expert~\cite{lipman2023flow} (width 448, one block per language-model layer) whose gradients do not reach the language model. It is then fine-tuned on the same clipped mixture for 50{,}000 steps at batch 64 with MolmoAct2's post-training recipe: learning rates $10^{-5}$ for the language model, $5\times10^{-6}$ for the vision encoder and connector, and $5\times10^{-5}$ for the expert, with 8 flow times per chunk and the action tokens trained beside the expert. It is evaluated through the expert with 10 Euler steps, under the same protocols as the token runs.

\textbf{Real-time chunking.} Table~\ref{tab:rtc} compares, for the NAT~VLA, the difference between consecutive executed actions at chunk switches with and without RTC.

\begin{table}[H]
  \centering
  \caption{Difference between consecutive actions}
    \label{tab:rtc}
  \footnotesize
  \setlength{\tabcolsep}{5pt}
  \begin{threeparttable}
  \begin{tabular}{lccc}
    \toprule
     & \multicolumn{2}{c}{Across a chunk switch} & Within \\
    \cmidrule(lr){2-3}
    NAT VLA & With RTC & Without RTC & one chunk \\
    \midrule
    Fine-tuned & $0.065 \pm 0.007$ & $0.126 \pm 0.010$ & $0.031 \pm 0.003$ \\
    Pretrained & $0.0096 \pm 0.0007$ & $0.046 \pm 0.009$ & $0.0024 \pm 0.0002$ \\
    \bottomrule
  \end{tabular}
  \tabnote{$|a_t - a_{t-1}|$ averaged over arm dimensions (in $[-1, 1]$), steps, and benchmarks; $\pm$: 95\% bootstrap interval. \emph{Across a chunk switch}: from the last executed action of a chunk to the first of the next; \emph{within one chunk}: two consecutive actions of the same chunk. Fine-tuned: LIBERO, RoboMimic, Meta-World; pretrained: held-out Bridge, RT-1, BC-Z, SO-100/101.}
  \end{threeparttable}
\end{table}